\documentclass[11pt]{article}
\usepackage{blurb_style}

\usepackage{natbib}
\usepackage{latexsym}
\usepackage{algorithm}
\usepackage{algorithmic}
\usepackage{amsmath}
\usepackage{tabularx}
\usepackage{dcolumn}

\usepackage{booktabs}
\usepackage{todonotes}
\usepackage{listings}
\usepackage{color}
\usepackage{hyperref}

\definecolor{dkgreen}{rgb}{0,0.6,0}
\definecolor{gray}{rgb}{0.5,0.5,0.5}
\definecolor{mauve}{rgb}{0.58,0,0.82}

\newcolumntype{d}[1]{D{.}{.}{#1}}

\newcommand{\idf}{\operatorname{IDF}}
\newcommand{\tf}{\operatorname{TF}}

\newcommand{\ti}{\operatorname{TI}}
\newcommand{\overbar}[1]{\mkern 1.1mu\overline{\mkern-1.5mu#1\mkern-2.5mu}\mkern 2.5mu}

\usepackage{color}

\title{Assessing Topic Model Relevance: Evaluation and Informative Priors}
\author{Angela Fan, Finale Doshi-Velez, Luke Miratrix \\ [0.5em]
Harvard University}

\begin{document}
\maketitle

\begin{abstract}
Latent Dirichlet Allocation (LDA) models trained without stopword removal often produce topics with high posterior probabilities on uninformative words, obscuring the underlying corpus content. 
Even when canonical stopwords are manually removed, uninformative words common in that corpus will still dominate the most probable words in a topic.
In this work, we first show how the standard topic quality measures of coherence and pointwise mutual information act counter-intuitively in the presence of common but irrelevant words, making it difficult to even quantitatively identify situations in which topics may be dominated by stopwords.
We propose an additional topic quality metric that targets the stopword problem, and show that it, unlike the standard measures, correctly correlates with human judgements of quality as defined by concentration of information-rich words.  
We also propose a simple-to-implement strategy for generating topics that are evaluated to be of much higher quality by both human assessment and our new metric. 
This approach, a collection of informative priors easily introduced into most LDA-style inference methods, automatically promotes terms with domain relevance and demotes domain-specific stop words.
We demonstrate this approach's effectiveness in three very different domains: Department of Labor accident reports, online health forum posts, and NIPS abstracts.
Overall we find that current practices thought to solve this problem do not do so adequately, and that our proposal offers a substantial improvement for those interested in interpreting their topics as objects in their own right.
\end{abstract}

\section{Introduction}

Latent Dirichlet Allocation (LDA) \citep{lda} is a popular model for
modeling topics in large textual corpora as probability vectors over
terms in the vocabulary. LDA posits that each document $d$ is a
mixture $\theta_d$ over $K$ topics, each topic $k$ is a mixture
$\beta_k$ over a common, set vocabulary of size $V$, and $w_{d,n}$, the $n^{th}$ word in
document $d$, is generated by first sampling a topic $z_{d,n}$ from $\theta_d$ and then drawing
a word from that topic:
\begin{eqnarray*}
\theta_d \sim \mathit{Dirichlet}( \alpha ) & \beta_k \sim \mathit{Dirichlet}( \eta ) \\ 
z_{d,n} \sim \mathit{Mult}( \theta_d ) & w_{d,n} \sim \mathit{Mult}( \beta_{z_{d,n}} ) .
\end{eqnarray*}
The $\alpha$ and $\eta$ are hyperparameters to be selected by the user.
Once such a model is fit, the $K$ topics are then commonly interpreted by looking at the most
probable words in their distributions $\beta_k, k=1,\ldots K$.
Unfortunately, \emph{stopwords}---words with no contextual
information---often dominate these lists of highest probability words.
Stopword-dominated topics are uninterpretable as semantic themes, and
even if canonical stopwords are removed, topics dominated by overly
general and uninformative words still reduce the utility, reliability,
and acceptance of statistical topic models by users outside of the
machine learning community.

To improve topic quality, practitioners typically rely on heavy pre-
and post-processing, such as creating stopword lists and re-training
the LDA models without those words.  Broadly, stopwords can be divided
into two categories: canonical (``the,'' ``and'') or domain-specific
(``child,'' ``son'' in a corpus about children). Canonical stopwords
can often be removed by referring to standard, publicaly available
lists.  Constructing lists of domain-specific stopwords, however, is a
non-trivial task and risks introducing human bias if the model trainer
builds these lists over repeated LDA runs. Such extensive processing
is also a challenge for scientific reproducibility, as typically many
preprocessing steps and deleted-word lists are not included in
publications. Further, many proposed automated or technical methods to improve topic quality
are complex and not easily integrated into existing software,
particularly for the applied LDA community or as part of a larger and
more complex graphical model.

In this work, we first expose a subtle but important concern regarding the evaluation of topic quality, as defined by concentration of information-rich words, when documents contain many irrelevant words: common metrics such as coherence \citep{lda_semantic_coherence} and pointwise mutual information (PMI)~\citep{topic_coherence_regularized_topic_models}, actually \emph{prefer} topics that place high probability on canonical stopwords.
Furthermore, these standard topic quality metrics cannot compare LDA models trained across different vocabularies, as is the
case when one is iteratively removing potential stopwords.
Worse, we demonstrate that, across several data sets and stopword removal schemes, these metrics do not appropriately correlate with human evaluations of interpretability in the presence of stopwords.

In sum, this work shows that (1) conventional approaches to the stopword problem for topic modeling are inadequate, and (2) this fact is possibly obscured because common measures of topic quality can be deceptive if the vocabulary used in the modeling is not heavily and carefully curated as a pre-processing step.
Not only do we demonstrate these two concerns, but we suggest a simple and easily implementable solution: use a heterogeneous collection of informative asymmetric priors on the topics to generate, in one model-fitting, different topics for the different words of interest. We show that this approach can both reduce the presence of stopwords and improve the domain relevance of topic models, assessed with human evaluation. We also provide an alternate evaluation metric, based on lift, that correlates well with human judgement of average word quality in topics to serve as a proxy when human evaluation is not feasible or practical. 
We suggest that lift be used in combination with other metrics to assess the many characteristics that contribute to high-quality topics.

\section{Related Work}

\paragraph{Metrics for Evaluating Topic Quality} 
The difficulties of measuring topic quality are well known. Traditional evaluation has used perplexity, but this has been shown to negatively correlate with human-measured topic interpretability using novel word and topic intrusion tasks (\citet{human_interp}). Since then, several other methods have been proposed to automatically evaluate topic quality. 
For instance, \citet{newman2010automatic} show that pointwise mutual information (PMI) correlates strongly with human assessments of semantic coherence. 
PMI measures the word association between pairs of topic words by using external data (often from English Wikipedia). 
\citet{lda_semantic_coherence} propose the topic coherence metric that measures topic word co-occurrence across documents to detect low quality topics, and show it correlates with expert topic annotations. 

However, previous work has noted that using single metrics to evaluate topic quality is problematic, as different metrics typically capture different facets of quality (\citet{roberts2014structural}).
For example, one might measure whether topics represent coherent, readable ideas, or measure coverage of the range of topics actually present in the corpus, or, as in our case, measure concentration of substantive, domain-relevant words in the final topics. 
In this work, we focus on the problem of stopwords, which present a modeling obstacle as they dominate the word frequency and co-occurrence statistics of a corpus. In many LDA models, topics mainly represent these common words, which obscures relevant corpus content. Further, we find that in the presence of stopwords, LDA metrics meant to evaluate other aspects of topic quality perform counterintuitively (see Section \ref{sec:theory}).

\paragraph{Methods for Increasing Topic Relevance and Reducing Stopwords}
To produce topics with more domain-relevant words, much work has
focused on automatic stopword detection and removal. Popular
techniques for automatic stopword detection include keyword expansion
and other information retrieval
approaches~\citep{stopwords1,stopwords2,stopwords3,hacohen2010initial}.
Several approaches identify stopwords based on term weighting
schemes \citep{ming2010vocabulary} or word occurrence distributions
\citep{wibisono2016dynamic,baradadcorpus}. 
\citet{mak} assume that every document has a type or label and
only include the words that are most correlated with the document
label while minimizing information loss. 
\citet{stopwords1} begin with a set of pre-generated search engine queries and quantify
word informativeness via the KL divergence between the query term
distribution and the corpus background distribution. These approaches
require parameter tuning to set various penalty
cutoffs. Several require document-specific labels
and/or query terms. In contrast, we propose a simple fix that can be
easily applied within existing LDA software frameworks.

More broadly, there are many efforts to improve the semantic
interpretability of topic models
\citep{mehrotra2013improving,yangefficient,xie2015incorporating,dirichlet_forest_priors,semantic_content_word_freq}. 
In particular, much work has improved topic quality via different
priors: \citet{lda_prior_setup} show the effectiveness
of general asymmetric priors to improve topic quality, \citet{topic_coherence_regularized_topic_models} use an
informative prior capturing short range dependencies between words,
and \citet{dirichlet_forest_priors} use Dirichlet
Forest priors to capture corpus structure. Other models modify LDA to
incorporate corpus-wide data of word frequency and exclusivity
\citep{semantic_content_word_freq,lexical_priors}, and focus on relative as well as absolute word frequencies in a topic. 

These modifications, however, are much more effective if overwhelmingly high-frequency words are removed first, as the majority of these models are not targeted towards isolating stopwords to improve topic readability. 
This stopword removal, needed to produce coherent output, is often conducted as a pre-processing step \citep{topic_coherence_regularized_topic_models,zhao2011comparing,jagarlamudi2012incorporating,dirichlet_forest_priors,lee2014low}.
Some models have some robustness towards the presence of
stopwords, but perform noticeably better with canonical stopword
deletion \citep{tan2010topic,zhao2011comparing}. Further, while many
methods have been proposed to identify stopwords and model only
domain-relevant words, many LDA users still extensively use canonical
stopword deletion, particularly for more complex graphical models, possibly because of the additional modeling burden of many of the methods above. 
Some work instead removes stopwords as a post-processing step rather than pre-processing \citet{schofield}, but this still necessitates a curated list of words to delete.

\section{Traditional topic quality metrics are not robust to stopwords.}   
\label{sec:theory}
We next show that two standard measures of topic
quality---coherence and PMI---perform counterintuitively in situations in which the
corpus contains many common but irrelevant words.  This situation is
common in many real corpora, where there is standard vocabulary that is
often repeated in the text but is generally uninformative.  That said,
our analysis does not invalidate the use of these measures in cases
where the vocabulary has been carefully curated for relevance.

After a discussion of coherence and PMI, we introduce another metric, log lift, that alleviates these found concerns in the case of the stopword problem.

\paragraph{Coherence} 
The \emph{Coherence} of topic $t$ is defined as
\[ \mbox{coherence}(t) := \sum_{i=1}^{M-1} \sum_{j=i+1}^M  \log \frac{D(v_i^t, v_j^t) + 1}{D(v_i^t)}, \]
where $v_i^t$ is the $i$th most probable word in topic
$t$ and $M$ represents the number of top topic words to evaluate.
$D(x)$ represents the number of documents word $x$ appears in, $D(x,
y)$ represents the number of documents $x$ and $y$ co-appear in, and
the $+1$ ensures the $\log$ is defined \citep{lda_semantic_coherence}.  Coherence is largest when
$D(v_i^t, v_j^t) = D(v_i^t)$, which occurs when either (a) the words
co-occur in a very small subset of documents and are absent elsewhere,
or (b) the words appear in all documents. The former case is
unlikely, particularly as topic evaluation is conducted on the top $M$
most probable topic words.  The latter case is not: it occurs when the
$M$ words evaluated are common words appearing in every document,
i.e. are stopwords.  For concreteness, in the ASD corpus, a corpus
about children with autism, ``Autism'' and ``child'' only co-occur in
3\% of documents, but ``and'' and ``the'' co-occur in 58\% of
documents.  The stopword ``the'' appears in 93\% of OSHA documents,
74\% of ASD documents, and 99\% of NIPS documents.  When averaged
across all topics, coherence is maximized when all top topic words are
common and overlapping.

\paragraph{PMI Score} 
The \emph{PMI Score} of topic $t$ is the median of
$\log p(v^t_i,v^t_j) / p(v^t_i) p(v^t_j)$ calculated for all pairs of the most probable words
$v^t_i, v^t_j$ within topic $t$, with $i, j \leq M$, where $p(x)$ is the
probability of seeing word $x$ in a random document, and $p(x, y)$ is
the joint probability of seeing x and y appearing together in a random
document.  These frequencies are traditionally estimated from text
outside of the corpus, e.g., from a snapshot of Wikipedia.  
The PMI for a pair $i,j$ of words is maximized if $v^t_i$ and $v^t_j$ co-occur. In
practice, this is easily achieved with high frequency words that
appear with high probability in all
documents---stopwords. Particularly in real world, noisy corpora,
domain words alone are relatively rare, so multiple domain-relevant
words co-occuring strongly is incredibly rare. For example, on the ASD
dataset, the words ``school'' and ``read,'' both fairly common words
in English and topics of high concern for parents, only co-occur in
1.5\% of documents. Variants of PMI, such as Normalized PMI
\citep{lau2014machine} adjust the frequencies to reflect specialized or
technical corpora but suffer similar drawbacks---they are still
maximized if topics are full of the same, high frequency, co-occurring
words.

\paragraph{An Alternate Measure: the Lift-Score}
In topic modeling, \emph{lift} \citep{taddy2012estimation} is the ratio of a
word's probability within a topic to its marginal corpus probability.
The lift of word $j$ in topic $t$ is defined as
$\frac{\beta_{tj}}{b_j}$, where $\beta_{tj}$ represents the
probability mass of word $j$ in topic $t$ and $b_j$ is the empirical
probability mass of word $j$ in the entire
corpus \citep{taddy2012estimation}.
Previous work has used lift to sort topic words, as it reduces the appearance of globally frequent terms
\citep{sievert2014ldavis}. 
We generate an overall topic quality metric by averaging the log lift
of the top 30 words of each topic.  This will ideally achieve two
ends: (1) if the top words in topics generally do not appear in other
topics, we will tend to find the topics to be well-separated and
distinct and (2) common words, such as stopwords, will tend to have
comparatively lower lift and so stopword-laden topics will have lower scores.  
Given this intuition, we expect this metric to better target the stopword aspect of topic quality. Our experiments show that lift is robust to the presence of stopwords, and we suggest that it can be used in combination with other LDA metrics for holistic topic evaluation.

\section{Empirical Evaluation of Metrics and Fitting Methods}
\label{sec:experiments}

In this section, we demonstrate the inadequacies that we mathematically argued regarding
traditional evaluation measures in Section~\ref{sec:theory} do indeed
present themselves on three datasets with varying characteristics. The
ASD corpus contains 656,972 posts from three online support
communities for autism patients and their caretakers. Posts contain
non-clinical medical vocabulary (e.g. ``potty going'' instead of
``toilet training'') and abbreviations (e.g. ``camhs'' for ``Child and
Adolescent Mental Health Services'').  The OSHA corpus contains 49,558
entries from the Department of Labor Occupational Safety and Health
database of casualties. 
Each entry describes a workplace accident.
Unlike the ASD corpus, the OSHA posts are
short and structured.  The NIPS corpus contains 403 abstracts from the
Neural Information Processing Systems Conference 2015 accepted
papers. These concisely written abstracts are of medium length with a
highly technical vocabulary and comparatively few traditional
stopwords.

Overall, for each corpus, we generate several collections of topics using several commonly used topic modeling methods from the literature for handling stopwords, as well as several variants of our new proposal designed specifically to accommodate and detect stopwords in a natural manner.
We first evaluate the collections for richness of substantive words as marked by experts and stopword rates.
We then see which quality metrics are appropriately associated with these metrics.
We finally see which underlying approach for topic generation is most successful.

\subsection{Topic Modeling Methods}  

We use several different baselines and several versions of our proposed approach.  
We describe these methods in the following sections.
The baselines we selected are extensively used by applied LDA users as well as the research community. 

\subsubsection{Basic Modeling Approaches} 
\label{sec:baselines}
We first consider three basic approaches to topic-modeling: (1) \emph{No
Deletion Baseline} --- standard LDA without stopword removal,
(2) \emph{Stopword Deletion Baseline} --- LDA deleting the 127
canonical stopwords from the Stanford Natural Language Toolkit, a
common preprocessing step and a standard canonical stopword list, and
(3) \emph{TF-IDF Deletion Baseline} --- LDA deleting words with TF-IDF
scores in the lowest 5\%, similar to the stopword removal work in \citet{stopwords1} and \citet{ming2010vocabulary}.
These, particularly canonical stopword deletion and TF-IDF based deletion, correspond to the approach many applied practitioners take.

We also have \emph{Hyperparameter Opt Baseline}, which is LDA with hyperparameter optimization as part of the LDA training.
Here researchers fit a series of models with different $\alpha$ and $\eta$, maximizing model fit and selecting the best fitting model, as measured by likelihood, as their final one. 
We use two versions of this baseline, one with the full vocabulary and one with canonical stopwords first deleted (the latter is most analogous to current state of practice).
This general approach is commonly thought to solve the stopword problem.
We will see, however, that our suggested priors can produce even more interpretable topics compared to optimizing these prior parameters with this approach.

\subsubsection{An Alternate Approach: Informative Priors} 
\label{sec:informative_priors}
We will see that the standard methods of practice outlined in Section~\ref{sec:baselines} can fail, and fail quite badly by having topic lists diminated by words deemed unuseful.
As an alternative we propose using an informative prior on $\eta$, which is just as simple to implement as stopword removal yet, as we will see in Section~\ref{sec:results}, yields more interpretable topics.  
The approach combines two ideas: (1) encourage the formation of different types of topics in the fitting process, in particular stopword topics and domain topics, by using different Dirichlet prior concentrations $\eta_t$ for different $t$ to model the corpus as a mixture of different types of topics that differently accommodate stopwords and domain-relevant words, and (2) for domain topics have an asymmetric prior $\eta_t$ that penalizes likely stopwords and promotes domain-specific words. 
Importantly, since we only need to change the prior concentrations $\eta_t$ to implement this approach, this approach can easily be used to augment more complex LDA extensions. 
This concept of domain-relevant and stopword topics is similar to \citet{paul2014discovering}, which proposed that there are two separate distributions that generate corpus words --- a background distribution that produces common words, and a foreground distribution that generates topical words. 

To understand the core idea of our informative prior approach, recall that the posterior distribution of a topic is essentially a mix of the emperical distribution of words thought to be members of the topic and the prior distribution assigned to that topic.
Typically this prior, a $V$-dimensional Dirichlet distribution, is symmetric with, say, all elements (weights) being 1, corresponding to a single pseudo-count for each possible word.
This prior regularizes the topics, pulling the posterior towards the prior mean.
If we increase the pseudocounts proportionally, we regularize more.
If we use asymmetric priors, then we will regularize towards the new prior mean defined by the normalized vector of weights.
We exploit both having different levels of regularization and having asymmetry in our proposed alternate priors we discuss next.

\paragraph{Stopword Topics ($\eta_0$)} 
The LDA model models all of the words in the document.
Thus, for good model fit, it is important that high-frequency words, even those with little information-bearing content, be explained somehow---we cannot just relegate them to low probabilities in all our
topics. 
Thus, of the $K$ topics, we let $I$ of them be stopword topics: $\beta_k \sim Dirichlet( \eta_o )$, where $\eta_0$ is uninformative $(1, 1, \dots, 1)$.  
This prior only mildly regularizes the word probabilities, allowing these explicit stopword topics to give high-frequency, but uninteresting, words a place to go.

\paragraph{Choices for Domain Topics ($\eta_1$): Term Weighting}
One intuitive prior-based penalization is to set the Dirchlet weights
$\eta_1$ for the domain topic priors to be the inverse of the
corpus unigram frequency, which gives high frequency words low prior
probabilities of occurring as a draw from a domain topic.  This
\emph{Word Frequency Prior} is our most na\"ive approach.
Even so, the overall model can achieve reasonable perplexities because frequent
corpus words can still be explained in the stopword topics. 
This sequestering of high frequency words to specific topics allows the
domain topics to more accurately reflect the nuances of the
corpus.

However, while penalizing words based on their frequency effectively
limits stopwords, it is not a targeted form of restriction --- it
equally penalizes a term that occurs a few times in many documents and
a term that occurs repeatedly in only a few documents (which is a signal of topic-relevance).  
We propose instead a
prior penalization for the domain-relevant topics proportional to the
TF-IDF score \citep{salton1991dat} of the word (again, the overall model can
achieve low perplexities because frequent terms can be explained by
stopword topics).  Our \emph{TF-IDF Prior} for $K$ topics in an LDA
model with $I$ stopword topics and $K-I$ non-relevant topics is
\begin{align*}
\beta_1, \ldots, \beta_I &\sim Dir(1, 1, \dots, 1) \\
\beta_{I+1}, \dots, \beta_K &\sim Dir \left(c_1 \overbar{\ti(w_1)}, \dots, c_1 \overbar{\ti(w_V)} \right)
\end{align*}
where $\overbar{\ti(w_v)}$ represents the average TF-IDF score of word
$v$ across the documents in the corpus and $c_1$ is an arbitrary scaling constant used to
appropriately size the TF-IDF scores.  
We use the common TF-IDF score
for word $v$ in document $d$ of $\tf(v, d)\log{\idf(v)}$.  This prior
shrinks the posterior probability of words with small TF-IDF scores,
e.g., common words that consistently appear across the corpus, in the
domain topics.

\paragraph{Choices for Domain Topics ($\eta_1$): Keyword Seeding}
The term weighting approach relies only on patterns of word usage within the
documents to create the prior.  However, in many situations, domain
experts may have additional knowledge about the corpus vocabulary that
the TF-IDF score does not take into account. In particular, many
domains have publicly available, curated lists such as key terms for
article abstracts, lists of medications, or categories of
accidents. We incorporate such information using \emph{keyword} topics
with a prior that reduces shrinkage on those pre-specified vocabulary
words. Similar to the TF-IDF prior, we set a $K$ topic LDA model to
have $I$ stopword topics and $J$ TF-IDF weighted topics, but then set
the remaining keyword topics to promote domain specific words:
\begin{align*}
\beta_1, \ldots, \beta_I &\sim Dir(1, 1, \dots, 1) \\
\beta_{I+1}, \dots, \beta_{I+J} &\sim Dir \left(c_1 \overbar{\ti(w_1)}, \dots, c_1 \overbar{\ti(w_V)} \right) \\
\beta_{I + J + 1}, \dots, \beta_K & \sim Dir(c_2 \gamma_1, \dots, c_2 \gamma_V) 
\end{align*}
where $\gamma_i = c$ with $c \gg 1$ if $w_i$ is a keyword and 1
otherwise, and $c_2$ is a scaling constant akin to $c_1$. The presence of the TF-IDF weighted topics serves a similar
purpose as the stopword topics --- providing topics for non-keywords
to fill discourages word intrusion into the keyword
topics. The large prior setting on relevant domain
keywords act as pseudocounts that counteract their lower corpus frequency compared to stopwords. 

We emphasize that these domain-specific keywords---words to
\emph{include} rather than words to exclude---are much distinct from
domain-specific stopword lists. The keyword lists used in our
experiments are large, generic, and downloaded off the internet. For example, for the ASD keywords, we take the entire list of symptoms and diseases from the Unified Medical Language System. We
find that these very general lists of domain terminology, when used as
keywords, significantly reduces the number of stopwords in those topics.
Generating lists of domain-specific terminology often does not require
an expert; it is easy to point other researchers to these sources of generic terminology
for reproducing experiments. In contrast, domain-specific
stopword lists---words to \emph{exclude}---often require, for a given corpus, an expert to engage in an iterative
process of pruning based on repeated LDA model runs.

\paragraph{Examined Variations}
In our evaluation we have (1) \emph{Word Frequency Prior}, consisting of a stopword topic and the rest word frequency prior topics, (2) \emph{TF-IDF Prior}, a stopword topic and the rest TF-IDF prior topics, (3) \emph{Keyword Seeding Prior}, a stopword topic, some TF-IDF topics, and some keyword seeding topics.  
We also have a \emph{Keyword Topics Baseline}, where we have all Keyword Topic priors with no predesignated stopword topic or TF-IDF topics.

\subsection{Evaluation Metrics}
Following standard practice, we take the $n$ most probable words in
each topic as the ones that define the topic.
We use $n=30$.
As our work is focused on reducing the effect of stopwords on topic quality, we consider two axes for evaluation: what proportion of top words were identifiable as stopwords and how many top words were identifiable as domain-relevant. We use both automatic and human evaluation.

To measure the number of stopwords, we report the percentage of
NLTK canonical stopwords.
To verify that the topics contained domain-relevant content, we asked
2 domain experts each in the medical autism and labor law domains to independently
identify terms deemed important to generate keyword whitelists. For
the NIPS corpus, we used the paper titles as whitelist words
(canonical stopwords were removed) under the assumption that titles are concise signals of content.  The average percentage of these
whitelist words in the top 30 words of each topic are reported as the
\emph{Expert Words}.  
This expert whitelist evaluation is related to the studies presented in \citet{human_interp}, which used Mechanical Turk to identify topic words that did not belong. 
We quantify the opposite---words pre-designated to belong by domain experts---for three main reasons. 
First, expert whitelists are a more
scalable evaluation method compared to Turk. Second, our corpora
require more specific domain knowledge for accurate topic evaluation,
making our topics less accessible for the average Turk worker.
Finally, unlike the generic keyword lists, the experts were very
selective in choosing important words from the corpora.  Thus, we also
report the co-occurence of the top topic words with the
expert-identified terms within documents (\emph{Codocument Appearance})
as a measure of whether our top words tend to co-occur with the
expert-produced lists.

\subsection{Parameter Settings}
For each dataset, we performed a grid-search over the number of topics (5 - 50 topics in
increments of 5), settings for prior weights $c_1$ (100, 10, 1,
$\frac{1}{10}$, $\frac{1}{100}$) and $c_2$ (100, 10, 1,
$\frac{1}{10}$, $\frac{1}{100}$), number of TF-IDF topics (1, 5, 10,
19), number of keyword seeding topics (1, 5, 10, 18, 19), weight of
keyword seeding $c$ (10, 50, 100, 1000), and number of Gibbs Sampling
iterations (100, 200, 500, 1000). 
Our models were largely insensitive to these choices: the number of stopwords and number of expert words deviated little. 
The most important parameter setting was that the total prior weight on the stopword topics ($\eta_0$) should be larger than the total prior weight placed on the TF-IDF topics. 
This encourages separation between stopword and domain topics by ensuring that stopwords are sufficiently penalized in domain topics.

We present results for $K=20$ topics, $c_1 = c_2 = 1$, and $c =
100$. Word Frequency Prior and TF-IDF Prior models were trained with
$I = 1$ stopword topic. Keyword Prior models were trained with $I = 1$
stopword topic, $9$ TF-IDF topics, and 10 keyword seeding topics.
For the ASD dataset, keywords for keyword seeding models were downloaded from the
Unified Medical Language System.  The keyword seeds used for the OSHA
corpus were pretagged in the dataset as one-word descriptors of the
accident (e.g. ``ship'' to indicate the accident occurred on a ship).
For NIPS, we used the list of 2015 NIPS submission category keywords.  
We emphasize that all of these keyword lists were produced
automatically, without any expert curation.

\begin{figure}[t]
\includegraphics[width=\textwidth]{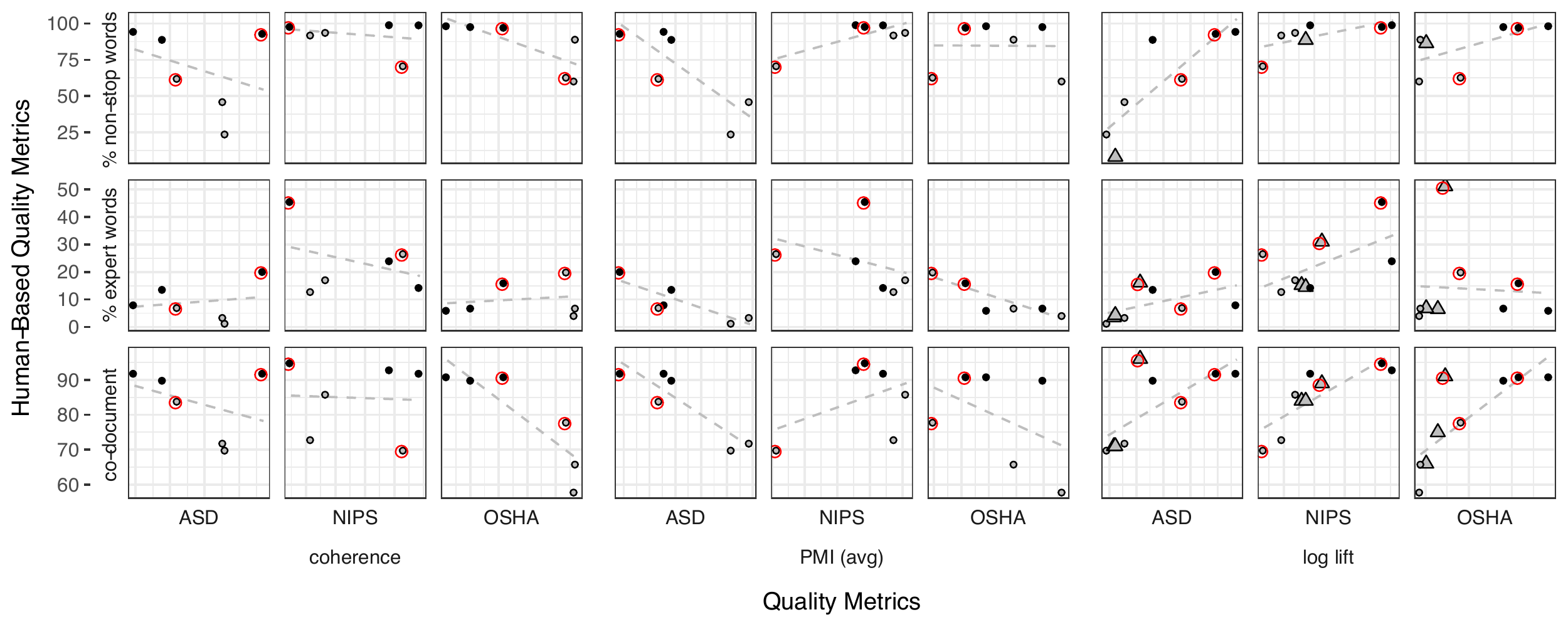}
\caption{Evaluation results on the top 30 words of each topic. Scatterplots of quality metrics with human-based quality metrics show log lift is generally correctly associated with human metrics but coherence and PMI are not. Solid points indicate prior-based approach, grey baseline. State-of-the-art baseline (both hyperparameter optimization baselines) and favored prior method (keyword prior) circled with red. Methods involving manual deletion marked with triangles. Methods with incomparable PMI or coherence due to differing vocabulary and methods with forced 0 stopword rate due to deletion are dropped.}
\label{fig:results}
\end{figure}

\begin{figure}[t]
\includegraphics[width=\textwidth]{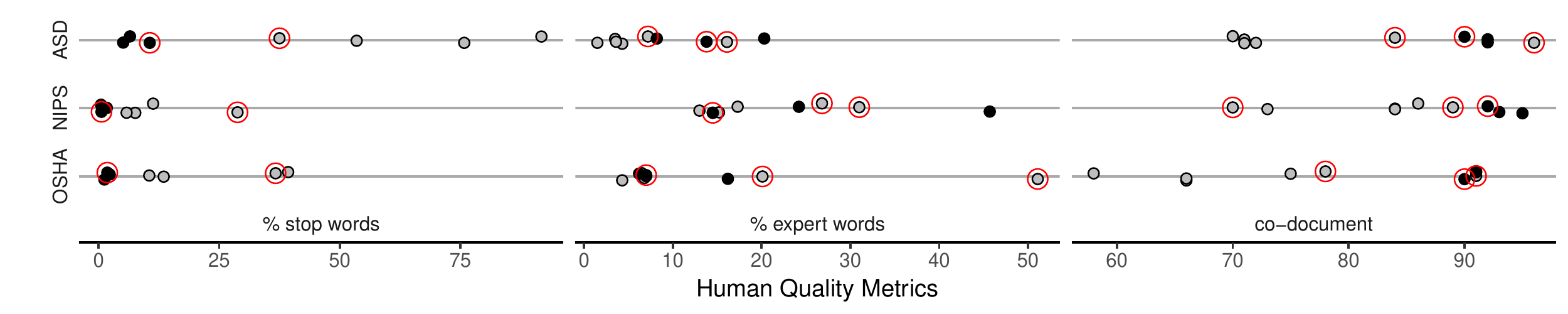}
\caption{Ranking of prior-based (black) vs. baseline (grey) approaches. See caption for Figure~\ref{fig:results} for details on further annotations. Generally the solid points (prior-based methods) are high for percent non-stopword (stop-word rate has been reversed to non-stop word rate), percent expert word, and co-document score, indicating superior performance.  Prior-based methods perform well, in general.}
\label{fig:results_ranking}
\end{figure}

\begin{table}[h!t]
\scriptsize
\centering
\begin{tabularx}{\textwidth}{rrX}
\multicolumn{3}{c}{\textbf{Qualitative Topic Evaluation}} \\
\toprule
& \textbf{Model} &  \multicolumn{1}{X}{\textbf{Topic}} \\ 
\midrule
\textbf{ASD} & No Deletion Baseline & social diagnosis as an or only are autism that child \\
& Stopword Deletion Baseline & schools lea information need special son statement parents support class \\
& TF-IDF Deletion Baseline & had just get school of will very not me out \\
& Keyword Topics Baseline & special lea it need has statement support needs school to \\
& Hyperparameter Opt Baseline & the to school needs support statement we permit chairman he\\
& Word Frequency Prior & reading paed attention helpful short communication cope aba diagnosed system \\
& \textbf{TF-IDF Prior} & \textbf{mobility improvements treatment preschool responsible expected friends panic professionals speak} \\
& \textbf{Keyword Seeding Prior} & \textbf{learning attention symptoms similar problem development negative disorder positive school} \\
\midrule
& \textbf{Example Stopword Topic} & child autism or on you it parent as son have \\

\midrule

\textbf{OSHA} & No Deletion Baseline & from approximately fell his hospitalized is him falling injured in \\
& Stopword Deletion Baseline & report trees surface backing inc degree determined forks fork board \\
& TF-IDF Deletion Baseline & his hospital due while work him in death pronounced tree \\
& Keyword Topics Baseline & head at injured falls for an balance fractures slipped lost \\
& Hyperparameter Opt Baseline & the employee lift number operator operating approximately jack to by\\ 
& Word Frequency Prior & mower limb top operator chain trees cutting log ground fell \\
& \textbf{TF-IDF Prior} & \textbf{collapses street trees lacerations wooden laceration construction chipper facility tree} \\
& \textbf{Keyword Seeding Prior} & \textbf{work rope tree landing protection caught lift edge open story} \\
\midrule
& \textbf{Example Stopword Topic} & hospitalized employee by for at when ft fall his fell \\

\midrule

\textbf{NIPS} & No Deletion Baseline & determinantal progress coordinate real learned cases theta ll super arms \\
& Stopword Deletion Baseline & includes top analytically margin framework incurs parameterizations normal confirmed ucsd \\
& TF-IDF Deletion Baseline & optimizations sgld finding others brownian strings logs generation recognize neurophysiological \\
& Keyword Topics Baseline & kappa keyword weights integrating similarly geometric dependence spatial definiteness either \\
& Hyperparameter Opt Baseline & the model learning bounded show algorithm algorithms feature optimization results\\
& Word Frequency Prior & mlfre vanilla wold validation inexactness benchmark gumbel bckw newton generalized \\
& \textbf{TF-IDF Prior} & \textbf{scalability variations index parameter parametric calibration versions condition infinite generalize} \\
& \textbf{Keyword Seeding Prior} & \textbf{nonlinear scalable newton optimisation hyperparameter stochastic optimality outliers epoch control}\\
\midrule
& \textbf{Example Stopword Topic} & based problem algorithms method from show be can learning data \\

\bottomrule
\end{tabularx}
\caption{\scriptsize Sample illustrative topics.  We present the top 10 words of a tree accident topic for OSHA, a school difficulty topic for ASD, and a hyperparameter tuning topic for NIPS. Informative prior model topics are more specific and contain fewer stopwords. Stopword topics from the informative prior models contain both domain specific and canonical stopwords.}
\label{tbl:qual_results}
\end{table}

\begin{table}[ht]
\footnotesize
\centering
\begin{tabular}{llrrrrllr}
	 &  & \multicolumn{2}{c}{Coherence} 	& Avg. & 	log	& \%  & \% &  Co-Doc \\ 
 Corpus & Model & 10 wds  & 30 wds & PMI 	& Lift &  Stopword &  expert  & Appear. \\ 
  \hline
ASD & No Deletion Baseline & -45.5 & -554.2 & -1.56 & 1.94 & 76\% & 2\% & 70\% \\ 
   & Stopword Deletion Baseline &  &  &  & 2.17 & 0\% & 4\% & 71\% \\ 
   & TF-IDF Deletion Baseline &  &  &  & 2.22 & 92\% & 4\% & 71\% \\ 
   & Keyword Topics Baseline & -48.2 & -580.1 & -1.42 & 2.61 & 54\% & 4\% & 72\% \\ 
   & Deletion + Hyp. Opt. &  &  &  & 3.13 & 0\% & 16\% & 96\% \\ 
   & Hyperparameter Opt. & -105.8 & -1107.9 & -2.12 & 4.73 & 38\% & 7\% & 84\% \\ 
   & Word Frequency Prior & -115.2 & -1278.3 & -2.02 & 3.65 & 15\% (11\%) & 14\% (14\%) & 90\% \\ 
   & TF-IDF Prior & -143.3 & -1611.8 & -2.08 & 6.71 & 10\% (5\%) & 9\% (8\%) & 92\% \\ 
   & Keyword Seeding Prior & -102.8 & -119.6 & -2.42 & 5.98 & 9\% (6\%) & 20\% (20\%) & 92\% \\ 
   \hline
NIPS & No Deletion Baseline & -71.2 & -790.7 & -2.06 & 2.96 & 8\% & 13\% & 73\% \\ 
   & Stopword Deletion Baseline &  &  &  & 3.58 & 0\% & 15\% & 84\% \\ 
   & TF-IDF Deletion Baseline &  &  &  & 3.72 & 11\% & 14\% & 84\% \\ 
   & Keyword Topics Baseline & -71.0 & -765.2 & -1.97 & 3.42 & 6\% & 17\% & 86\% \\ 
   & Deletion + Hyp. Opt. &  &  &  & 4.25 & 0\% & 31\% & 89\% \\ 
   & Hyperparameter Opt. & -72.7 & -633.2 & -2.96 & 2.35 & 29\% & 27\% & 70\% \\ 
   & Word Frequency Prior & -76.5 & -606.5 & -2.14 & 3.91 & 3\% (1\%) & 16\% (14\%) & 92\% \\ 
   & TF-IDF Prior & -86.7 & -656.8 & -2.35 & 6.60 & 4\% (0\%) & 24\% (24\%) & 93\% \\ 
   & Keyword Seeding Prior & -87.1 & -825.7 & -2.28 & 6.27 & 3\% (2\%) & 48\% (46\%) & 95\% \\ 
   \hline
OSHA & No Deletion Baseline & -68.2 & -831.9 & -2.66 & 2.89 & 39\% & 4\% & 58\% \\ 
   & Stopword Deletion Baseline &  &  &  & 3.29 & 0\% & 6\% & 75\% \\ 
   & TF-IDF Deletion Baseline &  &  &  & 3.02 & 14\% & 7\% & 66\% \\ 
   & Keyword Topics Baseline & -68.5 & -819.9 & -3.01 & 2.91 & 10\% & 7\% & 66\% \\ 
   & Deletion + Hyp. Opt. &  &  &  & 3.46 & 0\% & 51\% & 91\% \\ 
   & Hyperparameter Opt. & -74.8 & -899.1 & -3.60 & 3.85 & 37\% & 20\% & 78\% \\ 
   & Word Frequency Prior & -154.4 & -1738.6 & -2.80 & 4.83 & 6\% (2\%) & 8\% (7\%) & 90\% \\ 
   & TF-IDF Prior & -171.9 & -1951.2 & -3.21 & 5.87 & 5\% (1\%) & 7\% (6\%) & 91\% \\ 
   & Keyword Seeding Prior & -129.8 & -1447.4 & -3.36 & 5.18 & 5\% (2\%) & 17\% (16\%) & 91\% \\ 
     \end{tabular}
\caption{Full Table of Results. First columns are quality metrics with average topic coherence and average pointwise mutual information (closer to 0 is better) and average log lift calculated on the top 30 words of all models (large is good). Remaining columns are percent stopwords and percent content words in the top topic words, and co-document appearance of marked content words and top topic words in the documents. Numbers in parenthesis are percentages for domain topics only. We omit coherence and PMI for models with canonical stopword removal as they are not comparable due to different vocabulary sets.}
\label{tab:full_results}
\end{table}

\subsection{Results}
\label{sec:results}

Figure~\ref{fig:results} shows our primary results.
The top row shows how the different quality metrics correlate with percent canonical stopwords.
A well-performing metric would be negatively correlated.
Coherence and PMI are generally positively correlated or relatively flat.
The second and bottom rows correspond to the percent of words marked by experts as relevant, and the co-document occurrence of these expert words, respectively.
Here we would expect positive associations, and we instead see the associations to Coherence and PMI to generally be negative or flat.
Overall, coherence and PMI---two standard quality metrics for LDA---do not correlate well with our human evaluations.
By contrast, log lift has consistent negative association with stopword rate, positive association with co-document occurrence, and, other than the OSHA set, positive association with the percent of expert words.
We next discuss our findings in more detail.

\paragraph{Traditional Topic Quality measures do not correctly correlate with human measures of quality}

As stated above, Figure~\ref{fig:results} shows that coherence and PMI---two standard quality metrics for LDA---do not correlate with our human evaluations when the presence of stopwords is not addressed.
In fact, as shown on Table~\ref{tab:full_results}, we generally see the coherence and PMI scores are highest for the No Deletion Baseline, even though over 50\% of the top words are canonical stopwords.\footnote{For numerical comparability, we have to leave out the Stopword Deletion Baseline and TF-IDF Deletion Baseline, as their vocabulary sizes differ from our other baselines and proposed informative prior models.}
Similarly, the Keyword Topics Baseline falsely appears to perform well, despite containing both more canonical stopwords and less domain words than the full informative prior models.  
These results confirm our mathematical analysis in Section~\ref{sec:theory}: our standard quality measures systematically produce counterintuitive results when faced with irrelevant words.

We stress that these issues are not solved by stopword deletion; they plague topic model evaluation even for stopword deletion models, as these scoring mechanisms inevitably prefer topics composed of common words and domain-specific stopwords. 
For example, for the ASD corpus the Hyperparameter Opt Baseline (circled grey dots) appears to be a worse model when only looking at Coherence and PMI metrics, but clearly produces better topics compared to the No Deletion Baseline (see, e.g., Table~\ref{tbl:qual_results}). 
Standard metrics, while sensible in the absence of stopwords, produce results that prefer stopword-laden topics, and do not correlate with our human evaluation studies or expert topic evaluation.

\paragraph{Informative priors have superior quantitative performance.}
Across the three datasets, the models with informative priors generally produce
topics with (1) more domain-specific keywords deemed important by
experts and (2) fewer stopwords.  
See Figure~\ref{fig:results_ranking}: generally the informative prior models have small stopword rates, high co-document ratings, and generally fair to good proportions of expertly marked words.
Further, as Table~\ref{tab:full_results} shows, these models outperform baselines with a hard trimming threshold such as the TF-IDF Deletion Baseline \citep{stopwords1,ming2010vocabulary}. 
The stopwords that remain in topics with informative priors are almost all present in the predesignated stopword topics.
Informative priors increase the number of expert-designated domain-relevant words even though those words were \emph{not} used for the keyword seeding; our keyword seeds came from large, generic online lists. Most domain content appears in the domain-relevant topics, with the pre-designated stopword topic containing very few expert words. 
Additionally, the co-occurrence scores reveal that topic words from the informative prior model correlate more strongly with the independently-produced expert keywords.

In contrast, the deletion-based methods reduced the number of
canonical stopwords present but fail to capture as much domain content
and do not remove domain-specific stopwords. 

In a separate study of the ASD corpus, we had human evaluators identify the number of low-information words in the produced topics for three of the models: No Deletion baseline, Stopword Deletion Baseline, and TF-IDF Prior. 
10 evaluators were presented with the task of identifying low-information words, each assessed 2 runs of each model. 
No examples were given in order to avoid priming the identification of canonical stopwords.
In this experiment, human evaluators marked 71\% of the words in the No Deletion Baseline as stopwords, 26\% of words in the Stopword Deletion Baseline as stopwords, but only \textbf{17\%} of words in the TF-IDF Prior model as stopwords.  
Furthermore, for this model the 19 domain topics contained only \textbf{13\%} of marked stopwords, again indicating that the pre-desigated stopword topic can effectively sequester stopwords and prevent stopword intrusion into the domain topics. 

These results emphasize that canonical stopword deletion does not create
topics that humans judge to be stopword-free. 
In contrast, the prior models can create more readable, domain-specific topics with no vocabulary removal.

Lastly, simply seeding keywords as a prior without having topic types
(Keyword Topics Baseline) is not effective at reducing the
stopword effect or generating domain relevant topics (Table~\ref{tab:full_results}). 
The combination of penalizing priors and topic types is required for interpretable topics.

\paragraph{Informative prior topics are more readable}
The informative prior models generate more interpretable topics (see
Table~\ref{tbl:qual_results}).  
For example, in the OSHA dataset the baseline topics were overly general (see, for example, domain-specific
stopwords such as ``report'' from phrases such as ``an accident report
was filed'' in the deletion baseline) while the informative prior
models captured greater specificity (for example, one topic on
tree-related accidents from the Word Frequency prior shows accidents
often occur when ``cutting'' ``log[s].'')  In the ASD corpus, the
informative prior models captured specific concerns about
``learning,'' ``reading,'' and ``mobility'' for ASD patients entering
primary education.  In contrast, the deletion-baseline topics included
the domain-specific stopwords ``son'' and ``parent'' and addressed
school concerns only vaguely. In the NIPS dataset, the more concise
writing and technical terminology allow the baseline models' topics to
contain far fewer stopwords. However, the topic words do not form a
coherent theme with each other. In contrast, the topics for the
Keyword Seeding Prior and the TF-IDF Prior are much clearer as a
grouping. For example, the words ``optimization,'' ``hyperparameter,''
and ``epoch'' reference tuning various model parameters.

The learned stopword topics capture both canonical and domain-specific
stopwords. In the OSHA case, we see the words ``employee,'' ``ft,''
and ``hospitalized,'' as well as ``by,'' ``for,'' and ``at.'' For ASD,
we see ``child,'' ``autism,'' ``son,'' and ``parent'' as
domain-specific stopwords. In the NIPS dataset, the words ``problem,''
``algorithms,'' ``method,'' ``data,'' and ``learning'' are
domain-specific stopwords.

\paragraph{The Lift-Score predicts quality topics}
As shown on the rightmost 9 panels of Figure~\ref{fig:results}, the lift-score correlates with the quantitative performance metrics.
The informative prior models perform better overall than all baselines, with the TF-IDF Prior and Keyword Seeding Prior the best. 
Unlike Coherence and PMI, lift can be calculated across LDA models of varying vocabulary sizes and is not easily maximized by topics full of frequent words.

\section{Discussion}
The problem of stopwords is systemic---while LDA has been empirically
useful, it often picks up on spurious word
co-occurences as a result of lingual structure. For example,
researchers may wish to model important nouns, but these are often
preceded by articles such as ``the.''  LDA's bag of words assumption
treats these co-occurences as important indicators of words that
appear together, allowing stopwords to have undue influence. Much prior work that has focused on improving the quality of topics does not focus on the presence of stopwords due to the widespread usage of canonical deletion methods. Our work surfaces the relevant concern that domain-specific stopwords and other high frequency words reduce topic quality, even when using techniques such as hyperparameter optimization.

We expose an important gap in topic quality evaluation --- even if deletion methods are used to remove generic stopwords, human evaluators still judge the resulting topics to contain large quantities of low-information words. 
Furthermore, traditional topic quality measures did not reveal these trends.  
Our proposed lift-score, however, correlates to both human stopword evaluation and domain expert topic assessment, and can be used to assess topic quality in the presence of stopwords. 
However, more generally, an important question is to define an appropriate constellation of metrics that capture different factors such as concentration, uniqueness, coherence, and relevance, all of which are relevant to evaluating topic quality. 
Previous work has indicated that using individual metrics alone struggle to capture holistic topic quality \citep{roberts2014structural}, suggesting instead an evaluation of multiple metrics together. 
We believe assessing the presence of stopwords, domain or otherwise, is an important part of this overall evaluation strategy.

We also showed that simply adding specific informed priors that penalize
uninteresting occurrence patterns and promote relevant words can create
more interpretable topics by reducing the presence of domain-specific and canonical stopwords. 
In particular, the TF-IDF informative prior model not only drastically reduces the number of canonical stopwords appearing in the top 30 words of each topic, but also curtails the number of general, low information words.
These informed priors are easily incorporated into existing software by simply changing the existing symmetric Dirichlet prior on the word-topic distribution to one of the proposed priors, with no other inference modifications. 
See the Appendix for code snippets demonstrating this.
This ease of approach is particularly noteworthy for the Keyword Seeding Prior, as many other LDA models that incorporate external information require custom inference methods that may not be accessible to all users. 
We also found that our prior parameter settings are also quite robust and require little modification.  
Despite the large structural differences between the corpora, the same parameters produced interpretable topics that performed well both quantitatively and qualitatively.

Interesting avenues for future work include assessing our lift-score
metric with regards to additional human evaluations. It would also be interesting
to see whether incorporating informative priors into much more complex
topic models, such as supervised LDA models with correlational and
time-varying structure, provides similar gains. 
Implementation-wise, the simplicity of setting priors is a strength.
Informative priors could be easily incorporated into these more complex works. 
In fact, in these scenarios, more elaborate modeling of word frequencies might render the larger
effort computationally infeasible, making prior-setting even more critical.  
More generally, it would be interesting to see whether these more interpretable topics show benefits in downstream prediction tasks.

\clearpage
\section*{Acknowledgements}

Hanna Wallach, for thoughts on the project and pushing us to consider hyperparameter optimization. Thanks also to Chirag Lakhani and Tim Miller (and others) for scoring topics and topic words.

\bibliographystyle{apsr}
\bibliography{informative_prior_lda.bib}

\section*{Sample Code}
\begin{lstlisting}
import numpy as np
import math
import pickle
from scipy import sparse

# functions used for cleaning, saving, and loading data

def basic_clean(data, remove_set=set()):
    # don't model spaces or newline characters
    remove_set.add('')
    remove_set.add(' ')
    remove_set.add('\n')
    texts = [[word for word in document.lower().split() if word not in remove_set] for document in data]
    return texts

def create_vocabulary(data):
    """
    data is a list of documents, where each document is a list of words
    
    returns vocab, a mapping of word to index
    """
    vocab = {}
    counter = 0
    for document in data:
        for word in document:
            if word not in vocab:
                vocab[word] = counter
                counter += 1
    return vocab

def make_repr(data, vocab):
    """
    data is a list of documents, where each document is a list of words
    vocab is a dictionary mapping each vocabulary word to an integer
    
    returns output_matrix that is nDocuments by nVocab, where M_ij represents
    the number of times word j appears in document i
    """
    output_matrix = np.zeros((len(data), len(vocab)))
    for doc_index, document in enumerate(data):
        for word in document:
            output_matrix[doc_index][vocab[word]] += 1
    return output_matrix

def save_data(output_matrix, output_path):
    """
    data is saved in scipy sparse matrix format as a pickle file
    """
    sparse_m = sparse.csr_matrix(m)
    with open(output_path + ".pkl", "w") as f:
        pickle.dump(sparse_m, f)


# functions used to generate informative priors

text = ["Alice was beginning to get very tired of sitting", \
       "by her sister on the bank ,", \
       "and of having nothing to do :", \
       "once or twice she had peeped into the book her sister was reading", \
       "but it had no pictures or conversations in it", \
       "' and what is the use of a book , '", \
       "thought Alice", \
       "' without pictures or conversations ? '"]

keywords = ['book', 'pictures', 'conversations']
vocab = sorted(list(set(" ".join(text).split())))

def prior_data(data, keywords):
    data_dict = {}    
    num_documents = float(len(data))
    num_words = 0
    for index, words in enumerate(data):
        words = words.split(" ")
        doc_length = float(len(words))
        num_words += doc_length
        for word in set(words):
            word_count = sum([word == i for i in words])
            if word not in data_dict:
                data_dict[word] = {"wordCount": 0, "tf": {}, "keyword": 0, "numDocAppearance": 0}
            data_dict[word]["wordCount"] += word_count
            data_dict[word]["tf"][index] = word_count / doc_length
            data_dict[word]["numDocAppearance"] += 1
        
    for word in keywords:
        data_dict[word]["keyword"] = 1
        
    for key in data_dict:
        data_dict[key]["wf"] = data_dict[key]["wordCount"] / num_words
        data_dict[key]["idf"] = math.log(num_documents / data_dict[key]["numDocAppearance"])
        data_dict[key]["tfidf"] = np.mean(data_dict[key]["tf"].values()) * data_dict[key]["idf"] 
        
    return data_dict
  
def build_prior(prior_data, vocab, num_stopword_topics=0, num_wf_topics=0, num_tf_topics=10, num_keyword_topics=0, c1=1, c2=10):    
    def build_stopword_topic():
        return [1.0 for _ in vocab]
    
    def build_wf_topic():
        return [1.0 / prior_data[word]["wf"] for word in vocab]
    
    def build_tfidf_topic():
        return [c1 * prior_data[word]["tfidf"] for word in vocab]
    
    def build_keyword_topic():
        return [c2 * prior_data[word]["keyword"] for word in vocab]
    
    prior = [build_stopword_topic() for i in range(num_stopword_topics)] + \
            [build_wf_topic() for i in range(num_wf_topics)] + \
            [build_tfidf_topic() for i in range(num_tf_topics)] + \
            [build_keyword_topic() for i in range(num_keyword_topics)]
    
    return prior
  
# example usage:
prior_statistics = prior_data(text, keywords)
model_prior = build_prior("wf", prior_statistics, vocab, 1, 1, 1, 1)
\end{lstlisting}

\end{document}